\documentclass[letterpaper]{article} 
\usepackage{aaai2027}  
\usepackage[hyphens]{url}  
\usepackage{graphicx} 
\usepackage{natbib}  
\usepackage{caption} 
\usepackage{algorithm}
\usepackage{algorithmic}

\usepackage{newfloat}
\usepackage{listings}
\DeclareCaptionStyle{ruled}{labelfont=normalfont,labelsep=colon,strut=off} 
\floatstyle{ruled}
\newfloat{listing}{tb}{lst}{}
\floatname{listing}{Listing}

\usepackage{booktabs}

\usepackage{amsmath,amsfonts,bm}

\def\eqref#1{equation~\ref{#1}}

\def\1{\bm{1}}

\DeclareMathAlphabet{\mathsfit}{\encodingdefault}{\sfdefault}{m}{sl}
\SetMathAlphabet{\mathsfit}{bold}{\encodingdefault}{\sfdefault}{bx}{n}

\def\gL{{\mathcal{L}}}

\newcommand{\E}{\mathbb{E}}

\usepackage[most]{tcolorbox}
\usepackage{multirow}
\usepackage{subcaption}
\usepackage{pifont}
\newcommand{\cmark}{\ding{51}}%
\newcommand{\xmark}{\ding{55}}%
\lstdefinelanguage{json}{
  basicstyle=\ttfamily\tiny,
  showstringspaces=false,
  breaklines=true,
  frame=single,
  numbers=left,
  numberstyle=\tiny,
  keywordstyle=\color{blue},
  stringstyle=\color{red},
  commentstyle=\color{gray},
  morestring=[b]",
}

\title{Reinforcement Learning-based Semi-supervised Knowledge Distillation with LLM-as-a-Judge
}
\author{
    Yiyang Shen\textsuperscript{\rm 1},
    Lifu Tu\textsuperscript{\rm 2}\corresponding\thanks{Work done prior to joining Oracle.},
    Weiran Wang\textsuperscript{\rm 1}\corresponding
}
\affiliations{
    \textsuperscript{\rm 1}University of Iowa
    \textsuperscript{\rm 2}Oracle AI 

    \{yiyang-shen,weiran-wang\}@uiowa.edu,
    lifu.tu@oracle.com
}

\begin{document}
\nocopyright
\maketitle

\begin{abstract}
Reinforcement Learning (RL) substantially improves the reasoning capabilities of language models, but most existing RL fine-tuning approaches rely entirely on ground-truth verifiable rewards and thus labeled datasets with verifiable answers. To overcome this, we propose a RL framework for reasoning distillation that leverages continuous, LLM-based rewards. Our method employs an efficient mechanism that computes a continuous CoT reward (CCR) directly from a single-token logit of a judge LLM, evaluating the student model's reasoning trajectory. This formulation provides an effective and scalable online training signal that can be applied to large volumes of unlabeled data. 
We demonstrate that, when paired with a strong judge, simply using CCR achieves performance comparable to that of ground-truth or pseudo-label verifiable rewards, and even surpasses them as the amount of unlabeled data increases.
Furthermore, we find that combining them in a semi-supervised setup is highly synergistic: verifiable rewards help stabilize the CCR, while CCR improves the generalizability of verifiable rewards to related tasks. We also provide a comprehensive empirical comparison of various reward sources across multiple model architectures and dataset sizes. Our results show that this semi-supervised approach consistently enhances mathematical reasoning, yielding an absolute improvement of 5-10\% on multiple reasoning tasks.
\end{abstract}

\section{Introduction}
\label{sec:intro}

Recent advances have shown that reinforcement learning (RL) can significantly improve reasoning capabilities of large language models (LLMs), in particular on mathematical reasoning tasks where accurate reward signals can be obtained from verifiable answers~\citep{sutton1999policy,schulman2017proximal,ouyang2022training,rafailov2023direct,trung2024reft,guo2025deepseekr1}. However, RL-based reasoning training methods rely on labeled datasets and automatic verification, and their applicability remains constrained in settings where ground truth is unavailable (e.g., LLM generated question-answer pair not verified by human experts) and abundant training signals are thus difficult to obtain.

Knowledge distillation (KD) transfers reasoning ability from large teacher models to smaller students~\citep{hinton2015distillingknowledgeneuralnetwork,kim-rush-2016-sequence}. Mainstream methods such as supervised fine-tuning (SFT) rely on students attempting to reproduce teacher's reasoning trajectories~\citep{gu2026minillmonpolicydistillationlarge,deng2024distillation}. To overcome the lack of labeled data, a strong LLM can generate synthetic problems and reasoning trajectories for both SFT and RL with verifiable rewards (RLVR)~\citep{guo2025deepseekr1}. While effective, SFT approaches require large volumes of teacher demonstrations and primarily optimize next-token prediction rather than the underlying reasoning behavior. Naturally, reinforcement learning allows the student to improve its own reasoning ability by receiving feedback from a stronger teacher.

To address the reliance on ground truth, a growing line of research explores learning from weaker or learned supervision signals, including human preferences or model-based feedback~\citep{ouyang2022training,bai2022training,bai2022constitutional,zhu2023principled,rafailov2023direct}. Concurrently, the paradigm of utilizing an ``LLM-as-a-Judge'' has rapidly gained traction. However, this paradigm has been predominantly restricted to post-hoc evaluation---using a capable LLM to assess and rank the quality of generations~\citep{zheng2023judging,wang2023pandalm}. While these LLM evaluators correlate highly with human judgment, their potential to serve as an active, automated reward signal during RL fine-tuning remains underexplored. Existing methods that attempt to incorporate model-based feedback into training often distill it offline into static preference datasets~\citep{bai2022constitutional, lee2024rlaif} or involve complex, separate reward modeling pipelines that scale poorly to the massive sampling demands of continuous RL fine-tuning~\citep{rafailov2023direct, yuan2024self}. Therefore, we ask the following question:

\begin{quote}
\textit{Can we effectively use a capable LLM to provide active, continuous reward signals for training student models on unlabeled data?}
\end{quote}

In this work, we propose a \textbf{semi-supervised RL method for mathematical reasoning distillation}. It uses both labeled and unlabeled data for small LM fine-tuning, where a judge LLM evaluates the full reasoning trajectory of each unlabeled example to provide a reward signal.

A key feature of our framework is an efficient reward formulation based on single-token knowledge distillation, which we refer to as continuous chain-of-thought reward (CCR). To avoid complex reward modeling pipelines, we restrict the LLM judge to produce a binary decision (``yes'' or ``no'') evaluating the student's reasoning trajectory, and the reward is computed directly from the logits of these two tokens. Unlike a hard binary reward, the continuous probability derived from its logits contains nuanced information about reasoning quality. Furthermore, requiring only a single forward step for reward makes large-scale online RL practical.

Beyond introducing this reward, we perform an empirical comparison of different reward formulations prior to tuning, establishing that our method correlates well with verifiable rewards. Through subsequent training experiments, we demonstrate that CCR provides diverse and effective training signals beyond mere correctness. We find CCR alone yields performance comparable to that of ground-truth or pseudo-label verifiable rewards and even surpasses them as the amount of unlabeled data increases. Across mathematical reasoning benchmarks, combining verifiable supervision on labeled data with LLM judge rewards on unlabeled data consistently yields the strongest performance. Verifiable rewards help stabilize CCR, while CCR improves generalizability to related tasks, collectively demonstrating the effectiveness of semi-supervised RL for reasoning problems.

Our contributions to reasoning knowledge distillation 
are summarized as follows.
\begin{enumerate}
    \item We propose a \textbf{single-token judge reward formulation} that efficiently distills reasoning behavior by providing continuous \textbf{online} reward signals from binary next-token probabilities on full reasoning trajectories, a departure from mainstream approaches performed on offline datasets.    
    
    \item We introduce a \textbf{semi-supervised RL framework} that combines verifiable reward on labeled data with customized reward formulations on unlabeled math problems, substantially improving reasoning performance.

    \item We provide a \textbf{comprehensive empirical evaluation and comparison of different reward sources}, demonstrating their applicability to reasoning distillation. On tiny models (<1B), our RL framework achieves a 5--10 point absolute improvement on mathematical reasoning tasks compared to using only verifiable rewards. Moreover, it yields even larger gains on out-of-domain evaluations, demonstrating improved robustness. On a larger model (6.7B), our method still provides an improvement of approximately 1.5 points, indicating that the proposed approach scales effectively to larger architectures. 
\end{enumerate}

\section{Method}
\label{sec:method}
Reinforcement Learning with Verifiable Rewards (RLVR) has gained prominence following its use in DeepSeek-R1~\citep{guo2025deepseekr1} for o1-style~\citep{openai2024learning} long chain-of-thought (CoT,~\citealp{wei2022chain}) reasoning. Yet, because RLVR's benefits scale with model size, smaller architectures experience only modest reasoning improvements~\citep{guo2025deepseekr1}.
In this work, we demonstrate that a reinforcement learning framework guided by a carefully designed reward system can effectively elicit latent knowledge from a large model to enhance the reasoning capabilities of a smaller counterpart.

\subsection{Problem setup}

A supervised dataset consists of (question, CoT, answer) tuples denoted as $(x, e, y)$, where the chain-of-thought (CoT)~\citep{wei2023chain} $e$ represents the intermediate reasoning process that leads to the correct answer $y$.
The language model provides conditional probabilities of the form $\pi_{\theta}(a|s_t)$ where $a$ is a token in the vocabulary, and $s_t$ denotes the ``state'' at step $t$. The state can be interpreted as the generation context, including all tokens in the question and all tokens generated so far, based on which the LM predicts the next token. Once the token $a_{t+1} \sim \pi_{\theta}(a|s_t)$ is generated, it is appended to the exist context to form $s_{t+1}=[s_t,a_{t+1}]$ for generating the next token. This process is repeated until the model generates an end-of-sentence token $<\!\!\text{eos}\!\!>$. The initial state is $s_0=x$. 

\subsection{Training Pipeline} 
\paragraph{Warm-up.}
Depending on the base model's initial capabilities and the target task format, an initial Supervised Fine-Tuning (SFT) phase may be necessary to equip the LLM with complex reasoning capabilities. 
For instance, our GSM8K experiments require an SFT warm-up, whereas our AIME experiments bypass this stage. 
During SFT, the model is trained to maximize the log-likelihood of CoT sequences.
 Denote by $e=[a_1,\dots,a_{L-1},a_L=<\!\!\text{eos}\!\!>]$ the CoT token sequence of length $L$. The SFT loss is defined as
$
     \gL_{\text{SFT}} (\theta) = - \E_{e\sim D} \left[ \sum_{t=1}^{L} \log (\pi_{\theta} (a_{t}|s_{t-1})) \right]
$.

We can boost the model's reasoning capability by having the model learn from its experience using reinforcement learning~\citep{sutton2018reinforcement}. We view $\pi_{\theta}(a|s_t)$ as a \emph{policy} which generates tokens, or ``actions'' in the RL terminology, which evolves the state $s_t$ until reaching the terminal state (when $<\!\!\text{eos}\!\!>$ is generated), at which point a \emph{raw} reward is given based on the state trajectory. All intermediate steps have a raw reward of $0$.

\paragraph{Proximal Policy Optimization (PPO,~\citealp{schulman2017proximal}).}
The goal of RL is to maximize the expected reward given the policy. To this end, we adopt the Proximal Policy Gradient (PPO) method as implemented by~\citet{trung2024reft} for RL fine-tuning of a smaller LM. Let $r(s_{t-1},a_{t},s_{t})$ be the \emph{raw} reward for generating $a_{t}$ at step $t$; the design of raw reward is detailed in the next subsection. For stability, PPO regularizes the learned policy to stay close to the reference model $\pi_{\text{ref}}$ (e.g., a model obtained by SFT warm-up) using a KL divergence penalty, and it employs a value model $V_{\phi} (\cdot)$ to predict the expected reward for each state. 

\paragraph{Group Relative Policy Optimization (GRPO,~\citealp{shao2024deepseekmath}).}
We also explore GRPO, a variant of PPO that eliminates the need for a value model by sampling a group of responses and computing relative advantages. The formal definitions for PPO and GRPO are provided in Appendix~\ref{sec:ppo_grpo}. 
While GRPO inherently demands higher computational cost due to the generation of multiple hypotheses per question (group size $G=8$ for GSM8K, and $G=7$ for AIME in our experiments), we provide an empirical comparison in Section~\ref{expts:gsm8k} demonstrating that both algorithms yield comparable results on our tasks. 
Throughout this study, we follow the method of choice in prior works for each respective dataset.

\subsection{Reward Design}

\paragraph{Existing Reward Formulations.}
The majority of previous works rely on verifiable rewards (VR), using binary or partial rewards~\citep{le2022coderl,trung2024reft} to mitigate sparse signals~\citep{riedmiller2018sparse,trott2019solvingsparse}. For instance, the reward can be defined as:
\begin{equation*}
    r(s_{t-1},a_t,s_t)=
    \begin{cases}
        0, & s_L=\emptyset \\
        0.1, & s_L\not= \tilde y\;\text{and}\;s_L\not= \emptyset \\
        1, & s_L=\tilde y,
    \end{cases}
\end{equation*}
where $\tilde y$ is the ground truth answer and $s_L$ is the final answer string. 
In a natural language CoT setup, the model receives a reward of 0 if the final answer cannot be extracted, 0.1 if it can be extracted but is incorrect, and 1 if the extracted answer is correct. 
In a programming language (programming of thought, PoT) setup~\citep{gao2023pal}, the reasoning process is represented as a Python program, which is more rigorous and has lower perplexity than NL. Accordingly, we assign a reward of 0 if the code fails to compile or times out, 0.1 if it executes successfully but yields an incorrect answer, and a reward of 1 when the output matches the ground truth. We denote the loss function associated with this reward as $\gL_{\text{VR}}$. 

\label{sec:rerank}
Additionally, pretrained rerankers (PRR), typically used to select best-of-$k$ inference results~\citep{uesato2022rerank}, output a correctness probability. We can repurpose this probability as a continuous reward on unlabeled data, with the associated loss denoted $\gL_{\text{PRR}}$. However, relying solely on VR requires extensive labeled data, while PRR necessitates training a separate classifier for each specific task.

\paragraph{Our Novel Continuous CoT Reward (CCR).}
To overcome these limitations and effectively leverage abundant unlabeled data, we propose the Continuous CoT Reward (CCR). Rather than using a teacher LLM to generate complete CoTs---which is computationally heavy and prone to reasoning errors---we deploy it strictly as an online evaluator of the student's reasoning trajectories. 

During RL fine-tuning, the student model's generated candidate CoT is appended to an evaluation prompt forcing the judge LLM to answer only ``Yes'' or ``No''. We extract the logits corresponding to these two tokens to compute a continuous scalar reward. This provides a nuanced ``degree of correctness'' that serves as a rich training signal, drawing inspiration from the future constraint satisfaction score~\citep{tu2024unlocking}.
Specifically, the score is calculated as:
\begin{equation}
\begin{aligned}
    s(p(\texttt{Yes}),p(\texttt{No})) = \sigma \left(u \right)\cdot\mathbb{I}_{\{\sigma(u)\geq a\}},\\ \text{where}\; u=\frac{\log p(\texttt{Yes})-\log p(\texttt{No})}{\tau}.
\end{aligned}
\end{equation}
$\tau\geq 0$ is a temperature parameter, $\sigma(u)=\frac{1}{1+\exp(-u)}$ is the logistic sigmoid function, and $a$ is a threshold so that scores smaller than $a$ will be set to 0.
We emphasize the novelty and efficiency of CCR: the reward depends solely on the LLM logits from \textit{a single inference step} without executing code or generating full trajectories, and it remains largely task-invariant with minimal prompt adjustments. 

\paragraph{Novel Semi-Supervised Training Objective.}
A key contribution of our work is formulating a novel semi-supervised RL framework that balances ground-truth supervision with scalable, unsupervised feedback. Verifiable rewards (VR) ensure reliable grounding on labeled data, while unsupervised rewards---such as CCR and PRR, though other reward formulations can readily be incorporated---provide rich, continuous signals on unlabeled data. The overall objective is an 
convex combination of these losses:
\begin{gather*}
\gL_{\text{overall}}=\lambda \cdot \gL_{\text{VR}} + \mu \cdot\gL_{\text{CCR}} + \rho \cdot \gL_{\text{PRR}}
\end{gather*}
where $\gL_{r}$ denotes the loss (e.g., PPO or GRPO) using reward $r$, with non-negative coefficients satisfying $\lambda + \mu + \rho = 1$.

\section{Related Work}

\paragraph{Knowledge Distillation.} Knowledge distillation (KD) transfers knowledge from a large teacher model to a smaller student model~\citep{hinton2015distillingknowledgeneuralnetwork}. Early work in NLP extends KD to sequence generation tasks by distilling sequence-level outputs~\citep{kim-rush-2016-sequence}. A common approach is to minimize the Kullback–Leibler (KL) divergence between teacher and student distributions, which has been widely adopted in subsequent studies~\citep{kim-rush-2016-sequence,gu2026minillmonpolicydistillationlarge,wu-etal-2025-rethinking}. Beyond likelihood-based objectives, \citet{tu-18,tu-etal-2020-engine} proposed training an inference network to match the teacher by minimizing its energy function. Recent advancements have shifted focus toward distilling complex reasoning capabilities from LLMs. For instance, \citet{deng2023implicit} explore a novel paradigm of \textit{implicit} chain-of-thought distillation, transferring the explicit, token-by-token reasoning steps of a teacher directly into the vertical hidden states of a student to eliminate the need for generating intermediate text. Building on this push for extreme model compression and efficiency, recent frameworks like CODI~\citep{shen2025codi} utilize self-distillation to compress natural language reasoning into a continuous latent space by aligning hidden states across tasks, successfully matching explicit reasoning performance while drastically reducing computational overhead. Recent research has integrated RL to further elicit and distill these complex reasoning behaviors, as seen in DeepSeek-R1~\citep{guo2025deepseekr1}. However, conventional knowledge distillation via supervised fine-tuning (SFT) is limited to surface-level imitation, failing to capture underlying reasoning processes and resulting in poor generalization. 

On-policy distillation (OPD)~\citep{agarwal2024onpolicy,gu2024minillm,zhao2026selfdistilled,he2026selfdistillationzeroselfrevisionturns} trains the student using its own rollouts, leveraging the teacher's per-token log-probabilities as a dense reward. In contrast, our method utilizes an online reward computed over the entire reasoning chain on unlabeled data, avoiding token-level supervision.

\paragraph{Reinforcement Learning.} Reinforcement learning (RL) has become a standard approach for aligning large language models (LLMs) with desired behaviors. Early work on reinforcement learning from human feedback (RLHF) relies on reward models trained from costly human annotations~\citep{ouyang2022training,bai2022training}. To reduce this dependence, subsequent methods introduce model-based feedback, where LLMs replace human labelers~\citep{bai2022constitutional,lee2024rlaif}. However, these approaches typically construct static preference datasets and train separate reward models, resulting in complex and computationally expensive pipelines. To simplify training, methods such as Direct Preference Optimization (DPO)~\citep{rafailov2023direct} and related techniques~\citep{zhu2023principled,yang2024rlcd} optimize policies directly from pairwise preferences, while 
Self-Rewarding Language Models~\citep{yuan2024self} iteratively generate their own preference data for DPO training.
Nonetheless, these approaches remain limited by offline preference data. More recent work explores richer reward signals beyond binary feedback~\citep{damani2026beyond,gunjal2026rar}, and integrates RL with distillation to transfer reasoning capabilities~\citep{guo2025deepseekr1}. Here, we propose a reinforcement learning–based knowledge distillation framework that leverages multiple reward signals derived from LLM judges. Our approach produces lightweight, real-time rewards from unlabeled data, reducing the need for offline preference datasets and ground-truth supervision while achieving strong performance on reasoning tasks.

\section{Pre-training Evaluation of Reward Signals}
\label{sec:comp_reward}
Optimal distillation requires a high-quality judge and reward formulation on reasoning trajectories. For fair comparison, we select baselines requiring minimal token generation.

\begin{table}[t]
\centering
\begin{tabular}{cc}
\begin{tabular}{@{}l@{\hspace{0.0\linewidth}}c@{\hspace{0.02\linewidth}}c@{}}
\toprule
\bf Dataset & \bf Train & \bf Test \\
\midrule
\multicolumn{3}{l}{\textbf{Supervised}} \\
GSM8K           & 7,356 & 1,319 \\
SVAMP           & 3,043 & 1,000 \\
\midrule
\multicolumn{3}{l}{\textbf{Unsupervised}} \\
GSM8K-aug       & 385.5K & ---\\ 
GSM-Plus       & 9,233 & ---\\
\midrule
\multicolumn{3}{l}{\textbf{Evaluation}} \\
GSM-Plus       & ---     & 9,233\\
GSM-Symbolic   &       &\\
\hspace{1em} –main       & --- &5,000\\
\hspace{1em} –p1       & --- &5,000\\
\hspace{1em} –p2       & --- &2,500\\
\bottomrule
\end{tabular}
&
\begin{tabular}{@{}l@{\hspace{0.0\linewidth}}c@{\hspace{0.01\linewidth}}c@{}}
\toprule
\bf Dataset & \bf Train & \bf Test \\
\midrule
\multicolumn{3}{l}{\textbf{Supervised}} \\
\multicolumn{3}{l}{AIME 1983-2023} \\
& 919 & --- \\
\midrule
\multicolumn{3}{l}{\textbf{Unsupervised}} \\
\multicolumn{3}{l}{DAPO-Math-17K} \\
&14,110&--- \\
\midrule
\multicolumn{3}{l}{\textbf{Evaluation}} \\
AIME 2024               & --- &30\\
AIME 2025               & --- &30\\
AIME 2026               & --- &30\\
\bottomrule
\end{tabular}
\end{tabular}
\caption{Datasets Statistics. \textbf{Left}: Grade School-level Math. \textbf{Right}: Competition-level Math. }
\label{tab:datasets}
\end{table}

\paragraph{Likert.} We use the reference-Likert approach~\cite{zheng2023likert}, which provides the judge LLM with both a reference response and a candidate generation, prompting it to output a discrete score (e.g., 1--10). A notable drawback of this method is it creates extra CoTs, which increases latency and the risk of parsing invalid scores. Its reference-free version, direct-Likert, has been empirically shown to have poor performance as a reward signal~\cite{gunjal2026rar}.

\paragraph{RaR.} This baseline adapts the implicit Rubrics-as-Rewards (RaR) framework~\cite{gunjal2026rar}. Originally, RaR requires the LLM to generate detailed rubrics composed of criteria and weights for \textit{every} question, incurring high computational cost. For fair comparison, we use GPT-5.6 to generate one static rubric for GSM and another for AIME (Appendix~\ref{apdx:prompts}). The judge LLM then implicitly aggregates these criteria to score each response (1--10).

\paragraph{Token Likelihood (LL).} This approach utilizes the average token likelihood of the student's CoT as evaluated by the teacher model, measuring whether the student's reasoning trajectory aligns with the teacher's distribution. However, this method imposes significant computational overhead, as it requires calculating logits across the entire generated trajectory. This makes calculating reward for complex reasoning tasks that elicit long reasoning chains expensive.

\subsection{Empirical Comparison}
\label{sec:rwd_comp}
We perform experiments on both grade-level math and competition-level math, using datasets listed in Table~\ref{tab:datasets}.

\begin{table}[t]
\centering
\begin{tabular}{@{}l@{\hspace{0.03\linewidth}}c@{\hspace{0.03\linewidth}}l@{\hspace{0.03\linewidth}}c@{\hspace{0.03\linewidth}}c@{\hspace{0.03\linewidth}}@{}}
\toprule
\textbf{Signal} &
\textbf{Label} &
\textbf{Judge} &
\textbf{GSM8K} &
\textbf{SVAMP} \\
\midrule
VR
    & \cmark
    & ---
    & 100.0 & 100.0 \\
\midrule

\multirow{3}{*}{Likert}
    && Qwen3-8B           & 86.5 & 87.5 \\
    &\cmark & Llama3.1-8B & 72.0 & 80.5 \\
    && Gemma3-12B          & 79.5 & 85.5 \\
\midrule
\multirow{3}{*}{RaR}
    && Qwen3-8B            & 74.5 & 82.0 \\
    &\xmark& Llama3.1-8B & 57.5 & 62.0 \\
    && Gemma3-12B          & 70.0 & 72.5 \\
\midrule

PRR
    &\xmark & Galactica-6.7B
    & 87.5 & 80.5 \\
\midrule
\multirow{3}{*}{LL}
    && Qwen3-8B            & 81.5 & 79.0 \\
    &\xmark& Llama3.1-8B & 81.0 & 79.5 \\
    && Gemma3-12B          & 49.5 & 64.0 \\
\midrule
\multirow{3}{*}{\textbf{CCR}}
    && \bf Qwen3-8B            & \bf 88.0 & \bf 93.0 \\
    &\xmark& Llama3.1-8B & 82.0 & 87.5 \\
    && Gemma3-12B          & 85.0 & 87.0 \\
\bottomrule
\end{tabular}
\caption{Judge quality on Grade School-level Math.}
\label{tab:gsm-judge-quality}
\end{table}

\begin{table}[t]
    \centering
    \begin{tabular}{@{}l cccc@{}}
    \toprule
    \textbf{Judge Candidate}
    & \textbf{CCR} & \textbf{LL} & \textbf{RaR} & \textbf{Likert} \\
    \midrule
    Qwen3-8B    & 91.0 & 93.6 & 89.7 & 97.4 \\
    Qwen3.5-9B  & 94.9 & 92.3 & 71.8 & 95.0 \\
    Gemma3-12B  & 84.6 & 37.2 & 61.5 & 96.2 \\
    \bottomrule
    \end{tabular}
    \caption{Judge quality on AIME. }
    \label{tab:aime-judge-quality}
\end{table}


To evaluate the quality of the unsupervised reward signals, we sample 200 questions from the GSM8K, SVAMP, and the aggregated AIME datasets. For each question, we generate a paired correct ($y_w$) and incorrect ($y_l$) reasoning trajectory using the initial student model. We then prompt the candidate judge models to evaluate these trajectories and compute the respective rewards. A judgment is considered correct if the reward assigned to the correct trajectory exceeds that of the incorrect trajectory, i.e., $r(x,y_w) > r(x,y_l)$.

As shown in Table~\ref{tab:gsm-judge-quality}, Qwen3-8B consistently outperforms alternative models across reward formulations on GSM8K chain-of-thought trajectories; we thus adopt it as our primary grade school math judge. Furthermore, CCR demonstrates competitive reward accuracy while requiring far less runtime than generation-based rewards. Although PRR achieves exceptional in-domain accuracy on GSM8K (where it was explicitly trained on Galactica-6.7B trajectories), its performance degrades out-of-distribution on SVAMP.

Table~\ref{tab:aime-judge-quality} presents results on AIME. Although reference-based Likert reward yields the highest accuracy, its reliance on ground-truth labels precludes its use in unsupervised training. Among reference-free methods, combining Qwen3.5-9B and CCR achieves the best balance of reward accuracy and minimal computational overhead, so we select it as the AIME judge.

In Appendix \ref{apdx:reward_runtime}, we show that CCR takes the least amount of time among different formulations to be computed.

\section{Reasoning KD on Grade School-level Math}
\label{sec:kd_gsm}

\paragraph{Supervised Datasets.}
GSM8K~\citep{cobbe2021gsm} and SVAMP~\citep{patel2021svamp} are widely adopted grade school-level math datasets for training and evaluating reasoning LMs. 
We use their labeled splits for SFT and RL finetuning. Across all our grade school-level math experiments, models are trained to generate executable Python programs as reasoning (Program of Thoughts, PoT).

\paragraph{Unsupervised Datasets.} 
We deploy our proposed CCR reward signal across two unlabeled datasets: GSM8K-aug~\citep{deng2024distillation} and GSM-Plus~\citep{li2024gsmplus}. GSM8K-aug consists of questions generated by GPT-4; however, because it lacks rigorous quality filtering, the provided answers are occasionally incorrect, making it an ideal testbed for unsupervised reward signals. Although GSM-Plus typically serves as an evaluation benchmark, we repurpose its questions for unsupervised training by strictly masking all ground-truth labels during the RL phase.\footnote{We obtained explicit permission from~\citeauthor{li2024gsmplus} to use GSM-Plus for training on this project.} We selected GSM-Plus because its expert-reviewed, perturbed problems offer the necessary volume and structural complexity to push the boundaries of the model's reasoning capabilities.

\paragraph{Evaluation Datasets.} Several challenging benchmarks, including GSM-IC~\citep{shi2023gsmic}, GSM-Plus~\citep{li2024gsmplus}, GSM-Symbolic~\citep{mirzadeh2025gsmsymbolic}, and GSM-$\infty$~\citep{zhou2025gsminf}, have been proposed to evaluate the robustness of reasoning LMs. In this study, we focus on GSM-Plus and GSM-Symbolic.
Of the 10,552 samples in GSM-Plus, we exclude 1,319 instances labeled as ``None'' because they lack a valid ground-truth answer. 
GSM-Symbolic perturbs numeric variables from the original GSM8K dataset via templated generation. To further escalate the difficulty, it appends one (p1) or two (p2) additional clauses to each question, thereby increasing context length and structural complexity. 

\paragraph{Models.} We evaluate our distillation framework on two small LMs: Galactica-125M~\citep{taylor2022galactica} and CodeGen-350M~\citep{nijkamp2023codegen}.
We use Qwen3-8B as the CCR generator and pretrained Galactica-6.7B reward model as the PRR generator.
While computational constraints required us to employ fewer training epochs and a smaller global batch size compared to \citet{trung2024reft}, we successfully reproduce their core RLVR performance baseline.
Full implementation details are provided in Appendix~\ref{apdx:gsm_imp}.

\begin{table}[t]
\centering
\begin{tabular}{@{}l@{\hspace{0.01\linewidth}}c@{\hspace{0.01\linewidth}}c@{\hspace{0.02\linewidth}}c@{\hspace{0.03\linewidth}}c@{\hspace{0.03\linewidth}}c@{\hspace{0.02\linewidth}}c@{}}
\toprule
\textbf{Setup}
& \textbf{GSM8K}\;
& \textbf{GSM-Plus}
& \multicolumn{3}{c}{\textbf{GSM-Symbolic}}
& \textbf{SVAMP} \\
\cmidrule(lr){4-6}
& & & main & p1 & p2 &\\
\midrule
\multicolumn{7}{l}{\textbf{Original GSM8K sup}} \\
SFT & 25.25 & 15.14 & 22.88 & 7.40 & 3.64 & 32.30 \\
V & 30.48 & 18.38 & 30.96 & 8.42 & 0.96 & 31.30 \\

\multicolumn{7}{l}{\textbf{Pseudo-labeled GSM8K-aug sup}} \\
Base 1 & 28.89 & 16.78 & 25.00 & 6.48 & 2.16 & 44.60 \\
Base 2 & 36.01 & \bf 26.64 & 37.20 & 14.58 & 1.96 & \bf 51.00 \\
\multicolumn{7}{l}{\textbf{GSM8K-aug unsup}} \\
C & 33.13 & 21.28 & 30.38 & 11.36 & 3.32 & 44.70 \\
P & 32.98 & 20.92 & 31.12 & 10.36 & 2.60 & 46.30 \\
C+P & \bf 40.71 & \underline{25.33} & \bf 37.74 & \underline{14.62}& 3.00 & \underline{50.70}\\
\multicolumn{7}{l}{\textbf{GSM8K sup + GSM8K-aug unsup}} \\
V+C & 35.10 & 22.14 & 34.08 & 12.76 & 3.32 & 40.70 \\
V+P & 39.27 & 25.04 & 36.86 & 12.84 & 1.44 & 43.30 \\
V+C+P & \underline{40.11} & 25.26 & \underline{37.50} & \bf 15.70 & 2.24 & 42.60 \\
\bottomrule
\end{tabular}
\caption{Results of Galactica-125M as measured by accuracy (\%) on GSM8K. V = VR, C = CCR, and P = PRR. We apply equal weights (0.5) when combining two rewards, and set $\lambda=0.5$, $\mu=0.25$, and $\rho=0.25$ for V+C+P.}
\label{tab:galactica_gsm_semi_rl}
\end{table}

\subsection{Results on GSM8K}
\label{expts:gsm8k}

All RL experiments initialize from an SFT warm-up model achieving 21.83\% accuracy. A fully converged SFT model reaches 25.25\% accuracy on GSM8K, slightly exceeding \citet{trung2024reft}'s 23.70\% baseline.


\paragraph{Comparison against offline pseudo-labeling baselines.}
We evaluate two strong baselines where GSM8K-aug labels are generated \textit{offline} by a teacher model and trained using conventional RLVR. In \textbf{Base 1}, we constrain our teacher to complete the answer in a maximum of 40 tokens without reasoning. 
In \textbf{Base 2}, we use the full CoT trajectories generated by GPT-4 as pseudo-labels. Because these labels are model-generated and lack human verification, they contain noise. While \textbf{Base 2} achieves a strong 36.01\% on GSM8K (outperforming the original supervised baseline), it falls short of our proposed rewards, highlighting the superiority of online continuous evaluation over static offline pseudo-labels.

\paragraph{Unsupervised reward signals surpass supervised baselines.}
A compelling result of our framework is the strong performance of fully unsupervised RL training. As shown in the ``GSM8K-aug unsup'' section of Table~\ref{tab:galactica_gsm_semi_rl}, utilizing CCR (C) or PRR (P) independently achieves 33.13\% and 32.98\% accuracy on GSM8K, respectively, without relying on any ground-truth labels during RL. Strikingly, both unsupervised methods outperform the supervised verifiable reward baseline (V) trained on the original labeled GSM8K dataset, which achieves only 30.48\%. This superiority is largely driven by the scalability of our unsupervised approach: because our reward formulation circumvents the need for ground-truth labels, we can leverage vastly larger sets of unlabeled questions (e.g., GSM8K-aug) than would be possible with strictly supervised datasets. When combining both unsupervised signals (C+P), the performance jumps significantly to 40.71\%, highlighting that CCR and PRR can provide synergistic and diverse training signals. This demonstrates that scaling rich reward signals over abundant unlabeled data may effectively substitute costly annotations.

\begin{table}[t]
\centering
\begin{tabular}{@{}c@{\hspace{0.04\linewidth}}c@{\hspace{0.04\linewidth}}c@{\hspace{0.04\linewidth}}c@{\hspace{0.04\linewidth}}c@{\hspace{0.04\linewidth}}c@{}}
\toprule
\textbf{GSM8K}\;
& \textbf{GSM-Plus}
& \multicolumn{3}{c}{\textbf{GSM-Symbolic}}
& \textbf{SVAMP} \\
\cmidrule(lr){3-5}
 & & main & p1 & p2 \\
\midrule
\multicolumn{6}{l}{\textbf{GSM8K sup + 5\% GSM8K-aug unsup}} \\
 36.54 & 22.85 & 31.96 & 10.24 & 0.96 & \bf 43.10 \\
 
\multicolumn{6}{l}{\textbf{GSM8K sup + 25\% GSM8K-aug unsup}} \\
 38.36 & 24.61 & 36.42 & 13.74 & 1.16 & 42.30 \\
 
\multicolumn{6}{l}{\textbf{GSM8K sup + 100\% GSM8K-aug unsup}} \\
 40.11 & 25.26 & 37.50 & \textbf{15.70} & 2.24 & 42.60 \\

\multicolumn{6}{l}{\textbf{GSM8K sup + GSM-Plus unsup}} \\
 \bf 44.66 &  \bf 30.67 & \bf 38.74  & 10.88 & 0.84 & 32.50 \\

\bottomrule
\end{tabular}
\caption{More and higher quality unsupervised data improves performance on GSM8K. $\lambda=0.5,\mu=0.25,\rho=0.25$.}
\label{tab:galactica_tune_aug}
\end{table}

\paragraph{Semi-supervised RL provides superior performance and training stability.}
As shown in the bottom section of Table~\ref{tab:galactica_gsm_semi_rl}, combining labeled and unlabeled data under a semi-supervised setup yields the strongest results across GSM evaluation benchmarks. Integrating all three reward signals (VR, CCR, and PRR) achieves 40.11\% accuracy on GSM8K, representing a nearly 10\% absolute gain over conventional supervised RLVR (30.48\%). 
Furthermore, semi-supervised training is more stable than unsupervised training alone: while unsupervised models can experience model collapse (where actual task accuracy plummets despite high proxy rewards) after reaching peak accuracy, semi-supervised models maintain steady reward progression throughout training (a trend even more pronounced in Section~\ref{sec:svamp} below). 
This demonstrates that anchoring rich, continuous CoT rewards with verifiable ground-truth supervision provides both synergistic learning signals and robust optimization dynamics.

\paragraph{Strong out-of-distribution generalization.}
Beyond in-domain performance on GSM8K, our proposed continuous reward formulations exhibit generalization to out-of-distribution (OOD) benchmarks, including GSM-Plus, GSM-Symbolic, and SVAMP. As shown in Table~\ref{tab:galactica_gsm_semi_rl}, the standard supervised verifiable reward (V) baseline struggles to generalize, yielding only marginal OOD improvements or even regressions (e.g., dropping from 32.30\% to 31.30\% on SVAMP) over the SFT model. In contrast, our unsupervised (C+P) and semi-supervised (V+C+P) methods demonstrate robustness. For instance, the fully unsupervised C+P model improves GSM-Plus accuracy from 15.14\% (SFT) to 25.33\%, and achieves a leap on SVAMP from 32.30\% to 50.70\%. This indicates that reward signals generated by teacher models mitigate overfitting to the training distribution and foster deeper mathematical reasoning capabilities that transfer across diverse problem structures.


\paragraph{Scaling both quantity and quality of unsupervised data improves performance.}
Table~\ref{tab:galactica_tune_aug} demonstrates that scaling the amount of unsupervised data consistently improves model performance. Furthermore, scaling data \textit{quality} proves equally impactful: although GSM-Plus contains significantly fewer entries (9,233) than GSM8K-aug (385,461), training on it yields highly competitive results.\footnote{GSM-Plus labels are strictly used for evaluation, not fine-tuning (a valid transductive setup).} This indicates that utilizing a high-quality, human-reviewed dataset provides immense value, confirming that optimizing both data quantity and data quality are highly effective strategies.

\begin{table}[t]
\centering
\begin{tabular}{@{}l@{\hspace{0.0\linewidth}}c@{\hspace{0.0\linewidth}}c@{\hspace{0.02\linewidth}}c@{\hspace{0.04\linewidth}}c@{\hspace{0.04\linewidth}}c@{\hspace{0.02\linewidth}}c@{}}
\toprule
\textbf{Setup}
& \textbf{GSM8K}\;
& \textbf{GSM-Plus}
& \multicolumn{3}{c}{\textbf{GSM-Symbolic}}
& \textbf{SVAMP} \\
\cmidrule(lr){4-6}
& & & main & p1 & p2 & \\
\midrule

\multicolumn{7}{l}{\textbf{Reported by~\citet{trung2024reft}}} \\
V & 68.9 & \\ 

\multicolumn{7}{l}{\textbf{Continuing from the above checkpoint}} \\
V & 69.1 & 52.0 & 64.2 & 54.8 & 23.4 & 69.4 \\
\; + voting & 71.7 & 53.3 & 67.4 & \textbf{57.9} & 27.0 & 70.8 \\
V+C & 70.4 & 53.4 & 66.3 & 53.3 & 25.7 & 74.2 \\
\; + voting & \textbf{72.0} & \textbf{55.5} & \textbf{68.1} & 57.3 & \textbf{28.4} & \textbf{74.9} \\
\bottomrule
\end{tabular}
\caption{Results of finetuning Galactica-6.7B on GSM8K.
Here $\lambda=0.5,\mu=0.5$ for V+C.}
\label{tab:galactica_67b_ppo}
\end{table}

\paragraph{Scalability to Larger Architectures.}
To verify that our reward framework generalizes to larger student architectures, we apply our semi-supervised method to a 6.7B parameter model. As detailed in Table~\ref{tab:galactica_67b_ppo}, online continuous rewards boost RL performance even when the student and judge models are of comparable scale. Initializing from the Galactica-6.7B checkpoint provided by~\citet{trung2024reft}, we compare conventional RLVR against our semi-supervised objective ($V+C$).\footnote{Full $V+C+P$ on 6.7B requires ZeRO-3 offloading, making training prohibitively slow; we thus omit PRR.} 
Although the baseline RLVR performance is competitive at this scale, incorporating our continuous CoT reward over augmented data yields a notable gain, improving GSM8K accuracy from 69.1\% to 70.4\%. Applying majority voting~\citep{wand2023majorityvoting} with 100 hypotheses further elevates this to 72.0\%. These findings confirm that our distillation framework remains effective for larger models by capitalizing on vast amounts of unlabeled data.

\begin{table}[t]
\centering
\begin{tabular}{@{}l@{}c@{}c@{}ccc@{}c@{}}
\toprule
\textbf{Setup}
& \textbf{GSM8K}\;
& \textbf{GSM-Plus}
& \multicolumn{3}{c}{\textbf{GSM-Symbolic}}
& \textbf{SVAMP} \\
\cmidrule(lr){4-6}
& & & main & p1 & p2 & \\
\midrule

\multicolumn{6}{l}{\textbf{Original GSM8K only}} \\
SFT & 22.97 & 14.07 & 23.40 & 8.34 & 2.96 & 28.20 \\
V& 26.91 & 16.35 & 29.90 & 10.54 & 4.16 & 34.30 \\

\multicolumn{6}{l}{\textbf{GSM8K sup + GSM8K-aug unsup}} \\
V+C & 31.01 & 19.11 & \textbf{33.86} & 11.54 & 4.72 & \textbf{38.50} \\
V+P & 33.13 & 19.84 & 31.62 & 13.68 & 3.60 & 36.80 \\
V+C+P & 31.31 & 19.63 & 30.38 & 10.74 & 3.56 & 35.90 \\

\multicolumn{6}{l}{\textbf{GSM8K sup + GSM-Plus unsup}} \\
V+C & 35.63 & 22.98 & 33.12 & {12.08} & 2.08 & 32.80 \\
V+P & 36.24 & 24.12 & 29.94 & 12.34 & 3.12 & 32.70 \\
V+C+P & \bf 37.00 & \bf 24.99 & 33.00 & \bf 14.50  & 3.08 & 29.90 \\

\bottomrule
\end{tabular}
\caption{Results of CodeGen-350M on GSM8K.}
\label{tab:codegen_gsm_results}
\end{table}

\paragraph{CodeGen-350M Results.}
In Table~\ref{tab:codegen_gsm_results}, we show that similar performance gains are observed using another small model, CodeGen-350M.
With semi-supervised training on GSM8K-aug, we obtain 33.13\% on GSM8K test set when combining VR with PRR, 
showing a substantial increase from 26.91\% which uses only VR on labeled data. With higher quality unsupervised data (GSM-Plus), the accuracy of this setting is further improved to 36.24\%.
Clearly, the pre-trained rerank model is effective for a wide range of model sizes and different model accuracies on this dataset.
Our best result 37\% is achieved by semi-supervised training on GSM-Plus with both CCR and PRR.
We note that using only CCR or PRR on unlabeled data may lead to the reward hacking phenomenon, causing model collapse. However, VR helps stabilize semi-supervised training.

\subsection{Results on SVAMP}
\label{sec:svamp}

We evaluate our framework on SVAMP using both small LMs, applying optimal hyperparameters from GSM8K.
As shown in Table~\ref{tab:combined_svamp_results}, our semi-supervised framework yields significant performance gains on SVAMP across both small LMs. For Galactica-125M, incorporating augmented data with all three rewards improves accuracy from 40.1\% (supervised RLVR) to 44.3\%. Likewise, for CodeGen-350M, accuracy increases from 41.4\% to 46.6\%. 
Notably, pure unsupervised training without VR suffers from the aforementioned model collapse (see Appendix~\ref{sec:hacking} for detailed analysis). Integrating VR resolves this collapse, demonstrating stability of the semi-supervised setup.
Finally, we note an asymmetric generalization effect: training on GSM8K transfers effectively to SVAMP, whereas training on SVAMP yields poor performance on GSM8K (results not shown).


\begin{table}[t]
\centering
\begin{tabular}{l c c}
\toprule
\textbf{Setup}
& \textbf{Galactica-125M}
& \textbf{CodeGen-350M} \\
\midrule
\multicolumn{3}{l}{\textbf{Original SVAMP only}} \\
SFT & 37.40 & 35.80 \\
V & 40.10 & 41.40 \\
\multicolumn{3}{l}{\textbf{GSM8K-aug unsup}} \\
C & 44.50 & $\times$ \\
P & $\times$ & $\times$ \\
\multicolumn{3}{l}{\textbf{SVAMP sup + GSM8K-aug unsup}} \\
V+C & \textbf{47.00} & 41.80 \\
V+P & 42.80 & 43.20 \\
V+C+P & 44.30 & \textbf{46.60} \\
\bottomrule
\end{tabular}
\caption{Results on SVAMP. $\times$ indicates model collapse. We apply equal weights (0.5) when combining two rewards, and set $\lambda=0.5$, $\mu=0.25$, and $\rho=0.25$ for V+C+P.}
\label{tab:combined_svamp_results}
\end{table}

\section{Reasoning KD on Competition-level Math} 

We show below that our framework generalizes to harder natural language problems and newer model families eliciting much more sophisticated CoTs. 
We collect American Invitational Mathematics Examination (AIME) problems from 1983 to 2023 as our primary supervised dataset, where solutions typically require CoTs on the order of 10K tokens. 
Because AIME answers are strictly positive integers, we pair it with the English subset of DAPO-Math-17K~\citep{yu2025dapo}. We select DAPO-Math-17K because it underwent rigorous LLM-as-a-judge filtering and shares the positive-integer answer format. 
We train student models to output pure natural language reasoning for AIME and DAPO-Math, as these competition-level problems necessitate complex, abstract conceptual planning. 
Based on findings in Section~\ref{sec:comp_reward}, we adopt Qwen3.5-9B as our judge model. 
We employ Qwen3-0.6B operating in thinking mode as the student model. Given the difficulty of AIME problems and the variance across reasoning paths, we report average@32 accuracy.


\paragraph{Results.} 
Our AIME results in Table~\ref{tab:qwen_aime} corroborate our earlier findings. 
First, purely supervised RLVR on limited labeled data yields mixed results, improving AIME '24 but degrading performance on AIME '25 and '26. 
Second, purely unsupervised training with CCR provides consistent gains that scale directly with the volume of unlabeled data. 
Finally, combining both in our semi-supervised framework (V+C) resolves these inconsistencies, achieving the strongest overall performance across all evaluation sets, highlighted by a 7.5\% absolute gain on AIME '24.


\begin{table}[t]
\centering
\begin{tabular}{l c c c}
\toprule
\textbf{Setup}
& \textbf{AIME'24}
& \textbf{AIME'25}
& \textbf{AIME'26} \\
\midrule

HF Ckpt & 9.2 & 16.5 & 11.5 \\

\multicolumn{4}{l}{\textbf{Supervised}} \\
V & 10.7 & 15.1 & 11.0 \\

\multicolumn{4}{l}{\textbf{Unsupervised (until '23)}} \\
C & 12.0 & 16.9 & 12.7 \\

\multicolumn{4}{l}{\textbf{Unsupervised (until '23 + DAPO)}} \\
C & 15.5 & 18.3 & 13.0 \\

\multicolumn{4}{l}{\textbf{Semi-supervised} ($\lambda=0.1,\mu=0.9$)} \\
V+C & \textbf{16.7} & \textbf{20.7} & \textbf{15.2} \\

\bottomrule
\end{tabular}
\caption{Avg@32 results of Qwen3-0.6B-Thinking on AIME.}
\label{tab:qwen_aime}
\end{table}
\section{Conclusions}

We introduced a semi-supervised reasoning distillation framework powered by our novel Continuous CoT Reward (CCR). 
By combining sparse verifiable rewards with this continuous, inference-efficient teacher signal, our approach effectively circumvents model collapse and reduces dependence on labeled data, achieving significant gains on complex reasoning tasks.
Future directions include integrating more advanced policy gradient algorithms~\citep{yu2025dapo} into our distillation framework, and extending our methodology to fine-tune emerging diffusion language models~\citep{lou2024discrete,ye2024diffusion,nie2025large,zekri2025sepo,zhu2025llada15}.

\cleardoublepage
\bibliography{aaai2027}

\cleardoublepage
\appendix
\section{Policy Gradient Algorithms for RL}
\label{sec:ppo_grpo}

\subsection{Proximal Policy Optimization (PPO)}
For stability, PPO regularizes the learned policy to stay close to the reference model $\pi_{\text{ref}}$ (e.g., a model obtained by SFT warm-up), and its total reward takes into account the KL divergence between the learned model and the reference model: 
\begin{align*}
    r_{\text{total}}(s_{t-1}, a_{t}, s_{t}) &= r(s_{t-1}, a_{t}, s_{t}) \\
    &- \beta \cdot \mathbb{D}_{\mathrm{KL}}\!\left( \pi_\theta(\cdot \mid s_{t-1}), \pi_{\text{ref}} (\cdot \mid s_{t-1}) \right).
\end{align*}
Additionally, PPO employs a value model $V_{\phi} (\cdot)$ to predict the expected reward for each state. Architecturally, $V_{\phi}$ shares most parameters with the policy $\pi_{\theta}$ by adding a value head atop the final hidden layer.
We perform Temporal Difference (TD) estimation of state values:
\begin{gather*}
    \hat{A}_t = \sum_{l=0}^{L-t} (\gamma \lambda)^l \delta_{t+l},
    \quad \text{where}\quad
    \delta_{t'} = - V_{\phi}(s_{t'}) +\\ r_{\text{total}}(s_{t'},a_{t'+1},s_{t'+1}) + \gamma V_{\phi}(s_{t'+1})
\end{gather*}
with the terminal state value $V_{\phi}(s_{L+1}):=0$, $\gamma\in (0, 1]$ is the discount factor for rewards, and $\lambda\in [0,1]$ is the discount factor for TD~\citep{schulman2018highdimensional}. The final estimated state value for $s_t$ is 
$\hat{R}_t=\hat{A}_t + V_{\phi} (s_t)$. 
The loss for step $t$ involves two terms, one for the expected reward and the other for value estimation:
\begin{align*}
    \mathcal{L}_{\text{policy}}^t (\theta)
    & = \begin{aligned}[t]
        &- \mathbb{E}_{e \sim \pi_{\theta_{\text{old}}}}
        \Bigg[
            \min \Biggl(
                 \frac{\pi_\theta(a_t \mid s_{t-1})}{\pi_{\theta_{\text{old}}}(a_t \mid s_{t-1})} \hat{A}_t, \\
                & \operatorname{clip} \left(
                    \frac{\pi_\theta(a_t \mid s_{t-1})}{\pi_{\theta_{\text{old}}}(a_t \mid s_{t-1})},
                    1 - \epsilon,
                    1 + \epsilon
                \right) \hat{A}_t
            \Biggr)
        \Bigg]
    \end{aligned} \\
    \mathcal{L}_{\text{value}}^t (\phi)
    & = \begin{aligned}[t]
        &\frac{1}{2} \mathbb{E}_{e \sim \pi_{\theta_{\text{old}}}}
        \Bigg[
            \max \Biggl(
                 \left\lVert V_\phi(s_t) - \hat{R}_t \right\rVert^2, \\
                 &\left\lVert
                    \operatorname{clip} \left(
                        \hat{R}_t - V_\phi(s_t),
                        \hat{A}_t - \epsilon,
                        \hat{A}_t + \epsilon
                    \right)
                \right\rVert^2
            \Biggr)
        \Bigg]
    \end{aligned}
\end{align*}
where $\pi_{\theta_\text{old}}$ and $V_{\phi_\text{old}}$ denote earlier models that are used to sample the ``rollout'' CoTs and compute $\hat{A}_t$ and $\hat{R}_t$.
The final RL loss is the weighted combination:
\begin{gather*}
    \gL_{\text{PPO}}(\theta,\phi) = \sum_{t=1}^L \left( \mathcal{L}_{\text{policy}}^t + \alpha \mathcal{L}_{\text{value}}^t \right)
\end{gather*}
where $\alpha\ge 0$ is the coefficient for the value objective.

\subsection{Group Relative Policy Optimization (GRPO)}
Due to the success of DeepSeek-R1~\citep{guo2025deepseekr1}, GRPO is currently the most popular RL algorithm for reasoning tasks, and one may wonder how it compares with PPO on our tasks. Here, we briefly review the algorithm. 

A key architectural divergence lies in the absence of a value module in GRPO. Unlike PPO, which relies on a critic to estimate the advantage and mitigate the variance of the policy gradient, GRPO eliminates this requirement.
Instead, GRPO samples a group of CoTs for the same question, denoted as $\{e^1,\dots,e^G\}$ where $e^i=[a_1^i,\dots,a_{|e^i|}^i]$, using the old policy model $\pi_{\theta_\text{old}}$. These trajectories receive raw rewards $\{r^1,\dots,r^G\}$ at their corresponding terminal states. Outcome supervision provides the normalized reward that is shared by all tokens within each CoT:
\begin{gather*}
\hat{A}^i = \frac{r^i - \text{mean}(r)}{\text{std}(r)}.  
\end{gather*}
GRPO optimizes the following objective:
\begin{gather*}
    \gL_{\text{GRPO}}(\theta) = \mathbb{E}_{\{e^i\}_{i=1}^G \sim \pi_{\theta_{\text{old}}} } 
    \frac{1}{G} \sum_{i=1}^G \frac{1}{\lvert e^i \rvert} \times \\
    \sum_{t=1}^{\lvert e^i \rvert}
    \Biggl\{
        \min \Biggl[
            \frac{\pi_{\theta}(a_t^i \mid s_{t-1}^i)}{\pi_{\theta_{\text{old}}}(a_t^i \mid s_{t-1}^i)}
            \,\hat{A}^i,\,\\
            \operatorname{clip}\Biggl(
                \frac{\pi_{\theta}(a_t^i \mid s_{t-1}^i)}{\pi_{\theta_{\text{old}}}(a_t^i \mid s_{t-1}^i)},
                1 - \varepsilon,\,
                1 + \varepsilon
            \Biggr) \hat{A}^i
        \Biggr]\\
        - \beta \mathbb{D}_{\mathrm{KL}}\bigl[\pi_{\theta} \Vert \pi_{\text{ref}}\bigr]
    \Biggr\}.
\end{gather*}

\begin{figure}[t]
    \centering
    \includegraphics[width=\linewidth]{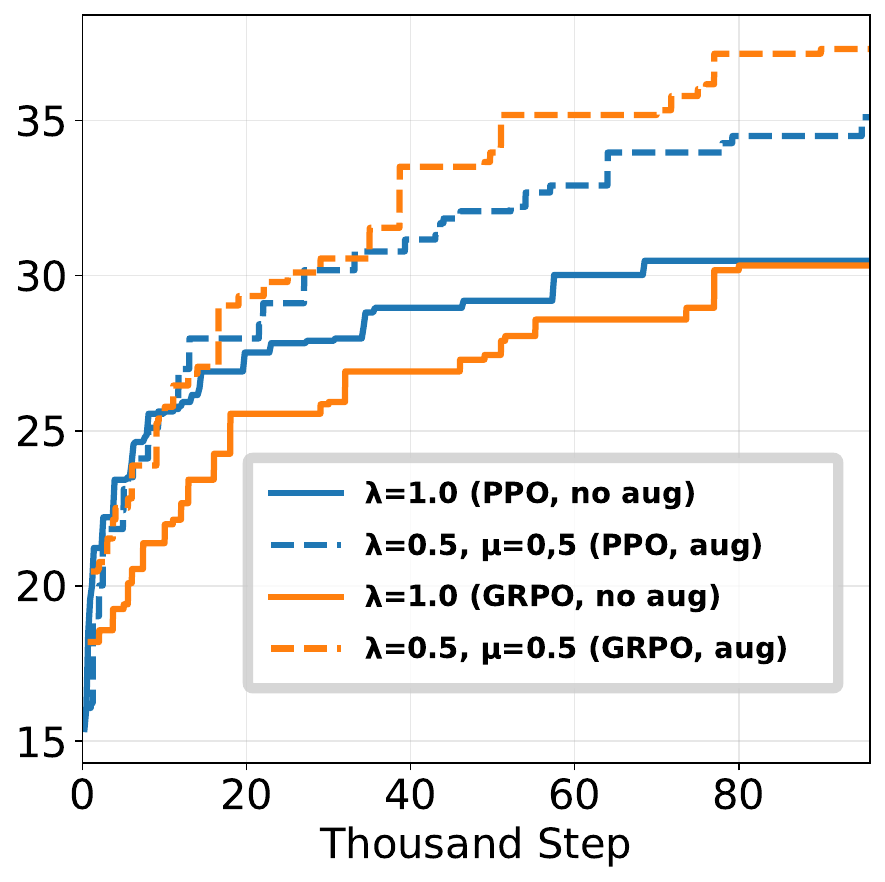}
    \caption{Performance comparison between PPO and GRPO. Learning curves for Galactica-125M trained on GSM8K. The V+C experiments use GSM8K-aug as the unsupervised dataset.}
    \label{fig:ppo_grpo_curve}
\end{figure}

\subsection{Comparison of PPO and GRPO}
Under the experimental setup described in Section~\ref{sec:kd_gsm}, we observe that GRPO achieves a larger performance improvement from semi-supervised training compared to PPO, as illustrated by the learning curves in Figure~\ref{fig:ppo_grpo_curve}. The final accuracies, summarized in Table~\ref{tab:galactica_125m_grpo}, demonstrate that GRPO more effectively leverages the unlabeled dataset (GSM8K-aug) in the V+C setup, yielding superior final performance across all evaluated benchmarks.

\begin{table*}[t]
\centering
\begin{tabular}{@{}lccccccc@{}}
\toprule
\textbf{Setup}\;
& \textbf{Algorithm}
& \textbf{GSM8K}\;
& \textbf{GSM-Plus}
& \multicolumn{3}{c}{\textbf{GSM-Symbolic}}
& \textbf{SVAMP} \\
\cmidrule(lr){5-7}
 & & & & main & p1 & p2& \\
\midrule
V & PPO & 30.48 & 18.38 & 30.96 & 8.42 & 0.96 & 31.30 \\
V+C & PPO & 35.10 & 22.14 & 34.08 & \textbf{12.76} & \textbf{3.32} & 40.70 \\
V & GRPO & 30.33 & 18.00 & 26.30 & 7.98 & 1.36 & 28.50 \\
V+C & GRPO & \textbf{37.30} & \textbf{23.71} & \textbf{35.36} & 11.18 & 2.20 & \textbf{45.20} \\
\bottomrule
\end{tabular}
\caption{Final accuracies obtained by PPO and GRPO for Galactica-125M. The V+C experiments use GSM8K-aug as the unsupervised dataset.}
\label{tab:galactica_125m_grpo}
\end{table*}

\section{Prompts and Templates}
\label{apdx:prompts}
In this section, we present all prompts used for training and evaluation.

\subsection{Task Templates for Student Model}
\paragraph{PoT for GSM problems} The SFT model completes the code based on the problem description provided in the docstring. 

\begin{tcolorbox}[
    colback=gray!10,
    colframe=black,
    boxrule=0.5pt,
    arc=2mm
]
\texttt{def solution():}

    \hspace{2em}"""\texttt{\{prompt\}}"""
\end{tcolorbox}


\paragraph{CoT for AIME problems}
We pass the following message directly to the model and extract the generated content within \verb|\\boxed{}| for verification.

\begin{tcolorbox}[
    colback=gray!10,
    colframe=black,
    boxrule=0.5pt,
    arc=2mm
]

\texttt{\{question\}}

\vspace{0.5em}

\textbf{Please reason step by step, and put your final answer within} \verb|\\boxed{}|.

\end{tcolorbox}

\subsection{Online Judge Reward Generation}
To compute the reward, we extract the judge LLM's specific logits for the ``Yes'' and ``No'' tokens from the output distribution at position $\Box$.
\paragraph{Evaluating Generated PoT on GSM Problems}
We instruct the judge LLM to evaluate candidate trajectories (both reasoning and final answer) using a prompt that requires a binary response. Note that the Python code is not executed, but rather statically evaluated by the LLM.

\begin{tcolorbox}[
    colback=gray!10,
    colframe=black,
    boxrule=0.5pt,
    arc=2mm
]

\textbf{You are a computer science teacher evaluating a student's Python solution.}

\vspace{0.5em}

\textbf{Question:} \texttt{\{question\}}

\vspace{0.5em}

\textbf{Python solution:} \texttt{\{response\}}

\vspace{0.75em}

\textbf{Ignoring the docstring, does the Python solution correctly solve the problem? Answer \textcolor{green!60!black}{\textbf{Yes}} or \textcolor{red}{\textbf{No}} only.}

\vspace{0.75em}

\textbf{Answer:} $\Box$

\end{tcolorbox}

\paragraph{Evaluating Generated CoT on GSM Problems}
We use the following prompt. The final answer string is included within the ``response'' block.
\begin{tcolorbox}[
    colback=gray!10,
    colframe=black,
    boxrule=0.5pt,
    arc=2mm
]

\textbf{You are a grade school math teacher trying to evaluate the student's response.}

\vspace{0.5em}

\textbf{Question:} \texttt{\{question\}}

\vspace{0.5em}

\textbf{Response:} \texttt{\{response\}}

\vspace{0.75em}

\textbf{Is the response correct? Answer \textcolor{green!60!black}{\textbf{Yes}} or \textcolor{red}{\textbf{No}} only.}

\vspace{0.75em}

\textbf{Answer:} $\Box$

\end{tcolorbox}

\paragraph{Evaluating Generated CoT on AIME Problems}
We use the following prompt. The final answer string is included within the ``response'' block.
\begin{tcolorbox}[
    colback=gray!10,
    colframe=black,
    boxrule=0.5pt,
    arc=2mm
]

\textbf{You are a college math professor trying to evaluate the student's response.}

\vspace{0.5em}

\textbf{Question:} \texttt{\{question\}}

\vspace{0.5em}

\textbf{Response:} \texttt{\{response\}}

\vspace{0.75em}

\textbf{Is the response correct? Answer \textcolor{green!60!black}{\textbf{Yes}} or \textcolor{red}{\textbf{No}} only.}

\vspace{0.75em}

\textbf{Answer:} $\Box$

\end{tcolorbox}

\subsection{Offline Pseudo-label Generation}

\paragraph{GSM8K-aug Base 1} The following prompt corresponds to Base 1 and is closely comparable to our method. Here, we instruct the teacher to generate the answer at position $\square$, which we decode and use as the pseudo-label.

\begin{tcolorbox}[
    colback=gray!10,
    colframe=black,
    boxrule=0.5pt,
    arc=2mm
]

\textbf{Solve the math word problem internally. Do not show reasoning.}

\textbf{Question: }\texttt{\{question\}}

\textbf{Answer this question with exactly one integer, decimal, or fraction. Answer:} $\square$
\end{tcolorbox}

\subsection{GPT-5.6 Generated RaR}
The generated Rubrics-as-Rewards (RaR) criteria for GSM and AIME are presented in Table~\ref{tab:gpt_rar_gsm} and Table~\ref{tab:gpt_rar_aime}, respectively.

\begin{table*}[htbp]
\centering
\begin{tcolorbox}[
    colback=gray!10,
    colframe=black,
    boxrule=0.5pt,
    arc=2mm
]
\begin{lstlisting}[basicstyle=\small\ttfamily, breaklines=true]
[
    {
        "title": "Correct Mathematical Result",
        "description": "Essential Criteria: Verifies that the Python code produces the correct final answer for the stated school math problem.",
        "weight": 5
    },
    {
        "title": "Valid Problem Setup",
        "description": "Essential Criteria: Checks that the code correctly translates all given quantities, operations, formulas, and constraints from the problem into Python.",
        "weight": 5
    },
    {
        "title": "Reasoning Before Conclusion",
        "description": "Important Criteria: Confirms that the response explains the mathematical reasoning or computational steps before presenting the final answer.",
        "weight": 4
    },
    {
        "title": "Code Reliability",
        "description": "Important Criteria: Checks that the Python code is syntactically valid, executable as written, and does not rely on undefined variables or unavailable data.",
        "weight": 4
    },
    {
        "title": "Clear Presentation",
        "description": "Optional Criteria: Rewards concise, readable code and explanations that use meaningful variable names and avoid unnecessary complexity.",
        "weight": 2
    },
    {
        "title": "Unsupported Hardcoding",
        "description": "Pitfall Criteria: Penalizes code that merely prints a guessed or preselected answer without performing calculations that correspond to the problem.",
        "weight": -2
    }
]
\end{lstlisting}
\end{tcolorbox}
\caption{GPT-5.6 Generated RaR criteria for GSM.}
\label{tab:gpt_rar_gsm}
\end{table*}

\begin{table*}[htbp]
\centering
\begin{tcolorbox}[
    colback=gray!10,
    colframe=black,
    boxrule=0.5pt,
    arc=2mm
]
\begin{lstlisting}[basicstyle=\small\ttfamily, breaklines=true]
[
    {
        "title": "Correct Answer",
        "description": "Essential Criteria: States a single final AIME answer as an integer from 0 to 999 that is mathematically correct for the problem.",
        "weight": 5
    },
    {
        "title": "Valid Reasoning",
        "description": "Essential Criteria: Uses logically sound mathematical reasoning in which each major step follows from previous steps without unjustified assumptions or invalid inferences.",
        "weight": 5
    },
    {
        "title": "Reasoning First",
        "description": "Important Criteria: Presents the solution process before giving the final answer so the conclusion is supported by the preceding argument.",
        "weight": 3
    },
    {
        "title": "Complete Solution",
        "description": "Important Criteria: Includes all key intermediate arguments or computations needed to justify the result rather than relying on unexplained leaps.",
        "weight": 4
    },
    {
        "title": "Clear Presentation",
        "description": "Optional Criteria: Organizes the solution clearly with readable notation and unambiguous mathematical statements that make the argument easy to follow.",
        "weight": 2
    },
    {
        "title": "Concise Writing",
        "description": "Optional Criteria: Avoids unnecessary repetition or irrelevant discussion while still providing sufficient justification for the solution.",
        "weight": 1
    },
    {
        "title": "Unsupported Claims",
        "description": "Pitfall Criteria: Makes unexplained assertions, skips essential logical steps, or presents a final answer without adequate mathematical justification.",
        "weight": -2
    }
]
\end{lstlisting}
\end{tcolorbox}
\caption{GPT-5.6 Generated RaR criteria for AIME.}
\label{tab:gpt_rar_aime}
\end{table*}
\cleardoublepage

\section{More Details on Different Reward Formulations}
\label{apdx:reward_runtime}

We use one NVIDIA RTX Pro 6000 GPU for every experiment. We sample 200 problems and solutions from Galactica-125M over GSM8K and SVAMP using the same random seed, respectively. For AIME, we sample solutions from Qwen3-0.6B, and select problems that have at least two correct and one incorrect answers. (We need two correct so that we have a reference answer for Likert score.) In total, there are 78 such AIME problems.

In Table \ref{tab:gsm-judge-runtime} and \ref{tab:aime-judge-runtime}, we compare the time it takes to obtain different reward signals and compare them in Section \ref{sec:comp_reward}. Our CCR method is the fastest as it requires single-token inference. PRR, while also fast, requires pretrained classifier model and may not be generalized to substantially different tasks.

\begin{table}[htbp]
\centering
\begin{tabular}{@{}lcc@{}}
\toprule
\textbf{Judge} & \textbf{GSM8K} & \textbf{SVAMP} \\
\midrule
\multicolumn{3}{@{}l}{\textbf{Likert}} \\
Qwen3-8B      & 360  & 204 \\
Llama3.1-8B   & 1098 & 587 \\
Gemma3-12B    & 760  & 456 \\
\midrule
\multicolumn{3}{@{}l}{\textbf{RaR}} \\
Qwen3-8B      & 152 & 196 \\
Llama3.1-8B   & 444 & 483 \\
Gemma3-12B    & 351 & 337 \\
\midrule
\multicolumn{3}{@{}l}{\textbf{PRR}} \\
Galactica-6.7B & 56 & 40 \\
\midrule
\multicolumn{3}{@{}l}{\textbf{LL}} \\
Qwen3-8B      & 93  & 67 \\
Llama3.1-8B   & 89  & 62 \\
Gemma3-12B    & 129 & 101 \\
\midrule
\multicolumn{3}{@{}l}{\textbf{CCR}} \\
Qwen3-8B      & 49 & 41 \\
Llama3.1-8B   & 53 & 38 \\
Gemma3-12B    & 82 & 68 \\
\bottomrule
\end{tabular}
\caption{Runtime (seconds) on Grade School-level Math.}
\label{tab:gsm-judge-runtime}
\end{table}

\begin{table}[htbp]
    \centering
    \begin{tabular}{@{}l cccc@{}}
    \toprule
    \textbf{Judge Candidate}
    & \textbf{CCR} & \textbf{LL} & \textbf{RaR} & \textbf{Likert} \\
    \midrule
    Qwen3-8B    & 16 & 23 & 68 & 82 \\
    Qwen3.5-9B  & 34 & 48 & 117 & 153 \\
    Gemma3-12B  & 29 & 36 & 119 & 162 \\
    \bottomrule
    \end{tabular}
    \caption{Runtime (seconds) on AIME.}
    \label{tab:aime-judge-runtime}
\end{table}

\section{Implementation Details for Grade School Math}
\label{apdx:gsm_imp}
Experiments for small LMs (Galactica-125M and CodeGen-350M) are performed on 4 NVIDIA RTX Pro 6000 GPUs.

\subsection{Data Processing}
\paragraph{Sampling.} We created 100 completions/responses from 200 randomly selected test questions, using the SFT warm-up model and same seed for each test set. We select one correct completion and one incorrect completion and use the LLM to score the correct completion and the incorrect completion.

\paragraph{Training Data.} The grade school-level math reasoning datasets (GSM8K and SVAMP) are directly obtained from the ReFT repository~\citep{trung2024reft}. The augmented version of GSM8K (GSM8K-aug) was originally created by~\citet{deng2024distillation}, and we utilize the processed version available on Hugging Face at \texttt{whynlp/gsm8k-aug}.
For our ablation study regarding the amount of augmented data, we randomly sample 5\% (19,273 examples) and 25\% (96,365 examples) of the full GSM8K-aug dataset.

\paragraph{Pseudo-label Base 1.} 
We include a baseline where the judge LM is used to generate pseudo-labels for GSM8K-aug, enabling us to train the RLVR baseline on this dataset (see Appendix~\ref{apdx:prompts} for the specific prompts and templates used). This generation process yielded 384,441 valid pseudo-labeled samples with 49.33\% of the completed answers matching the original labels of GSM8K-aug, while 1,020 invalid samples were discarded.

\subsection{Models}
All models are loaded from Hugging Face.
The student models are \texttt{facebook/galactica-125m} and \texttt{codegen-350M-mono}. The CCR judge model is \texttt{Qwen/Qwen3-8B}. The pretrained reranking model is \texttt{galactica-6.7b-ReFT-Rerank-GSM8k}.

\subsection{Hyperparameters}
\label{sec:gsm-hypers}

\paragraph{Training Steps.} Because we employ two distinct dataloaders for our distillation experiments (one for labeled data and one for unlabeled data), we track optimizer update steps rather than epochs. Furthermore, since the GSM8K-aug dataset is exceptionally large, a single epoch over the unlabeled data would take at least six days, making epoch counting impractical. 
For experiments trained on GSM8K, we trained Galactica-125M for 100,000 optimizer steps and CodeGen-350M for 50,000 steps. For the Galactica-6.7B experiments continuing from a checkpoint, we ran 15,000 optimizer steps. For all experiments on SVAMP, training proceeded for 50,000 steps. 

\paragraph{Supervised Fine-Tuning (SFT).} To warm-up a model, we train it for 10 epochs with a global batch size of 24. For complete SFT runs, we train the model for 40 epochs, also with a global batch size of 24. A learning rate of $2\times 10^{-5}$ is used in both scenarios.

\paragraph{RL-Based Distillation.} All small models are initialized from their warm-up checkpoints and trained using a global batch size of 24. The number of PPO epochs is set to 2. The value function coefficient $\alpha$ is set to 5 for Galactica-125M and 0.1 for CodeGen-350M. The KL coefficient $\beta$ is set to 0.01 across all Program-of-Thought (PoT) tasks.

\paragraph{CCR Reward.} We set $\tau=1$ and $a=0.35$.

\paragraph{GRPO.} We use a global batch size of 16 and sample 8 hypotheses ($G=8$) per question to calculate advantages. All other settings remain identical to those used in PPO.

\paragraph{Large Model Experiment.} We use a global batch size of 16 and sample 100 hypotheses per question for majority voting.

\paragraph{Overall Objective.}
While the reward functions could theoretically be combined into a single PPO loss, we maintain three separate losses since we draw from two different dataloaders (labeled and unlabeled). To ensure fair comparisons across methods, we record progress strictly in optimizer update steps rather than data-specific metrics.

\section{Additional Results on GSM}

\begin{figure*}[t]
\centering

\begin{minipage}{0.33\textwidth}
\centering
\includegraphics[width=\linewidth]{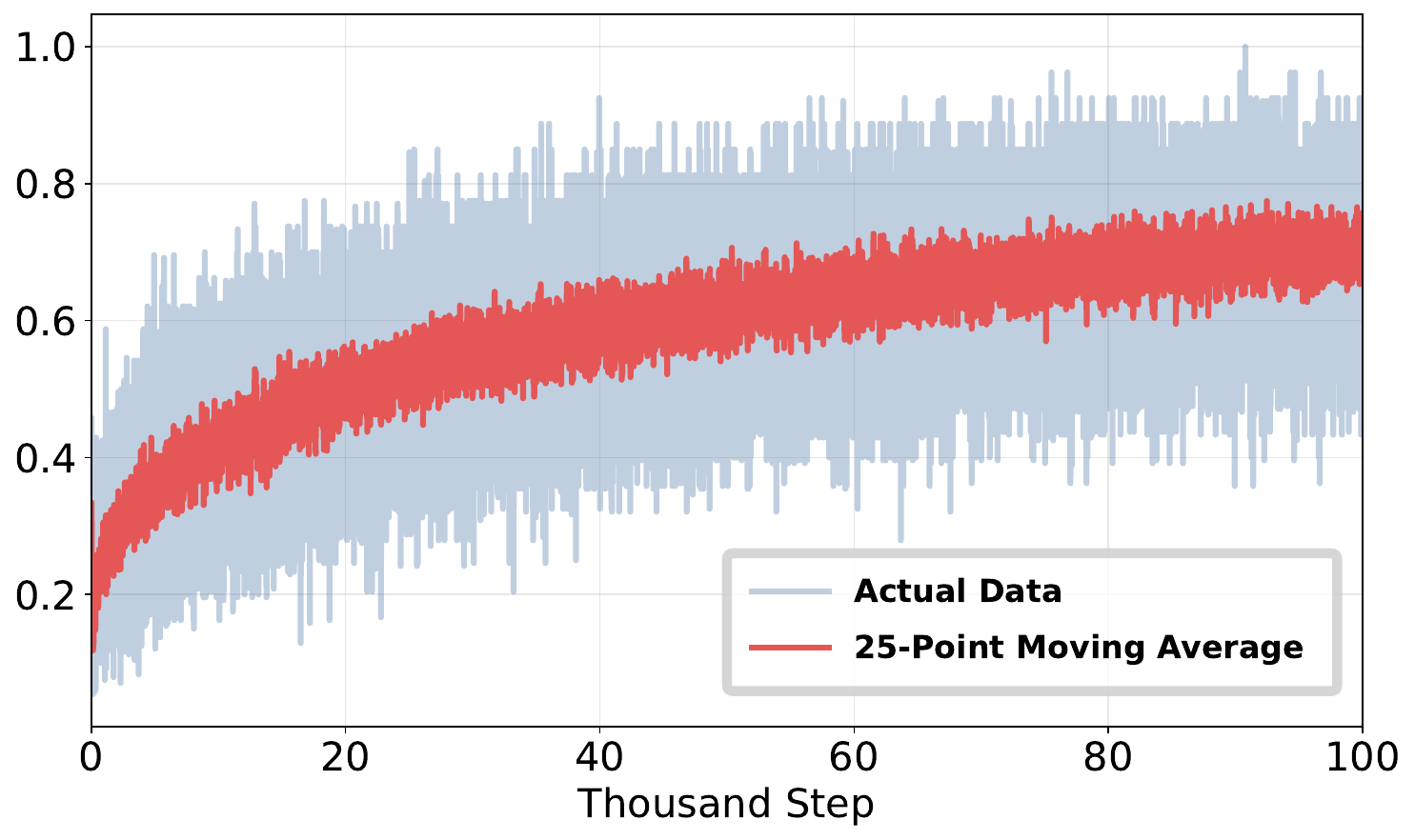}
\caption*{VR}
\end{minipage}
\hfill
\begin{minipage}{0.33\textwidth}
\centering
\includegraphics[width=\linewidth]{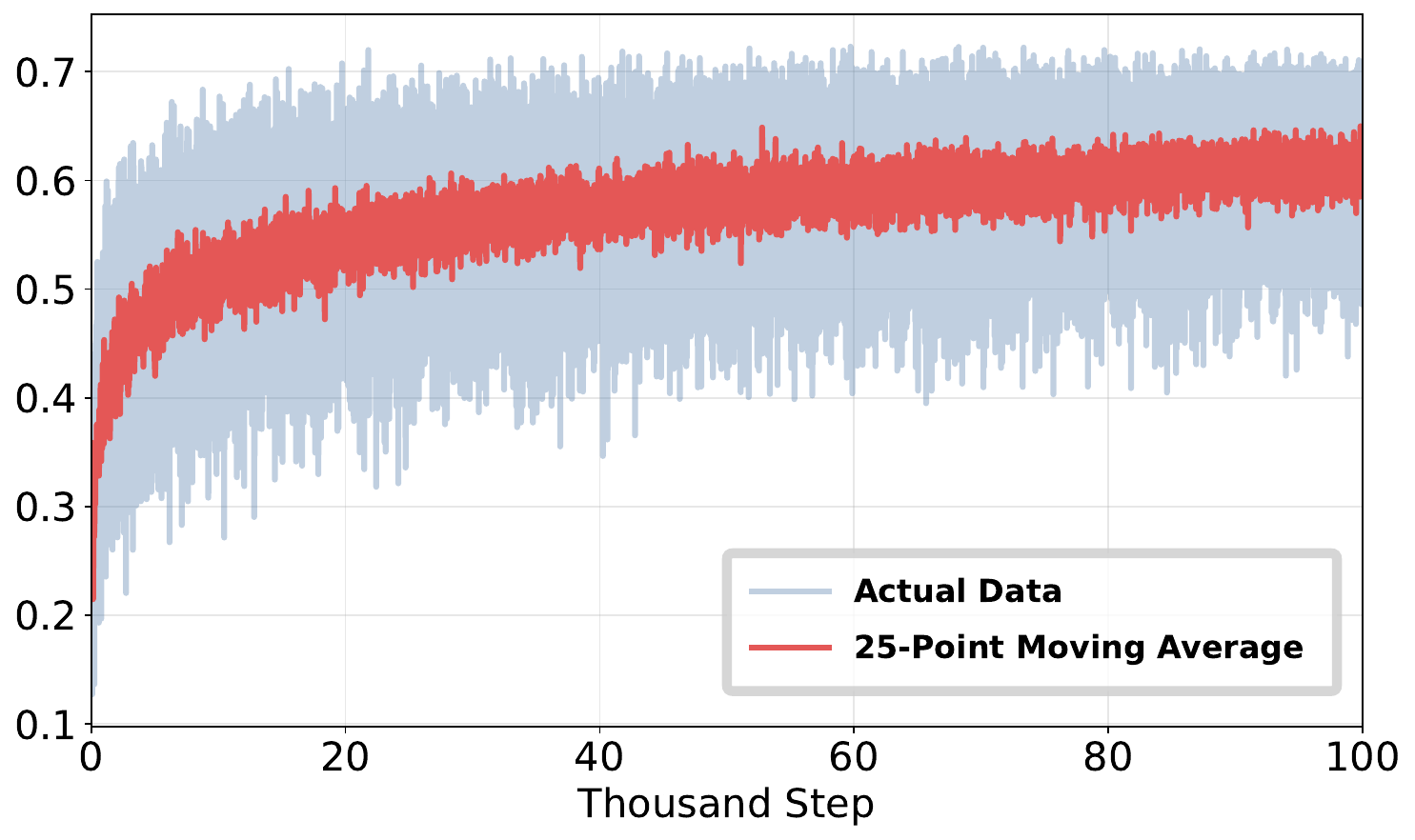}
\caption*{CCR}
\end{minipage}
\hfill
\begin{minipage}{0.33\textwidth}
\centering
\includegraphics[width=\linewidth]{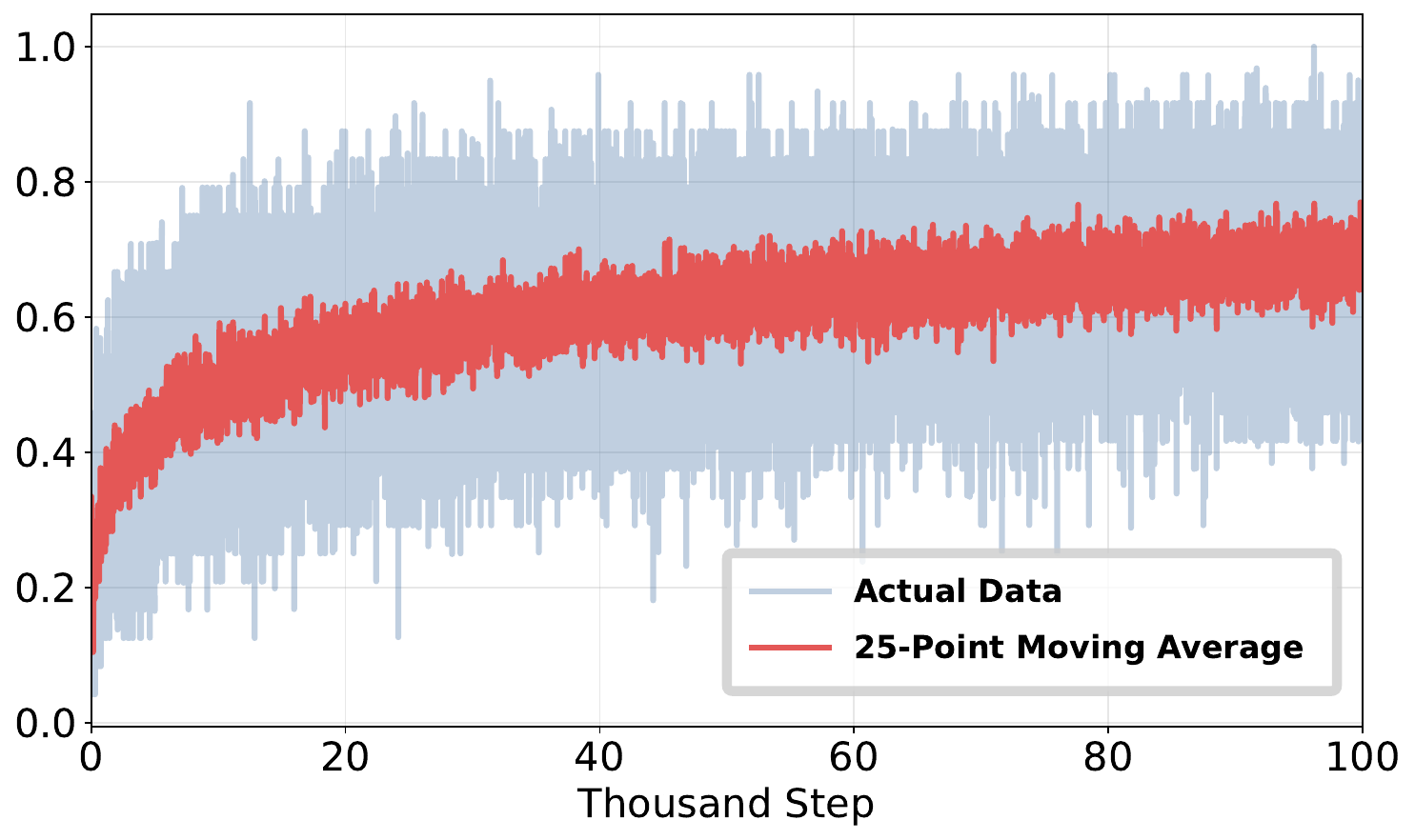}
\caption*{PRR}
\end{minipage}
\caption{All three types of our active training reward on Galactica-125M in a semi-supervised setup.}
\label{fig:reward_curves}
\end{figure*}

\begin{figure*}[htbp]
  \centering
    \includegraphics[width=0.5\linewidth]{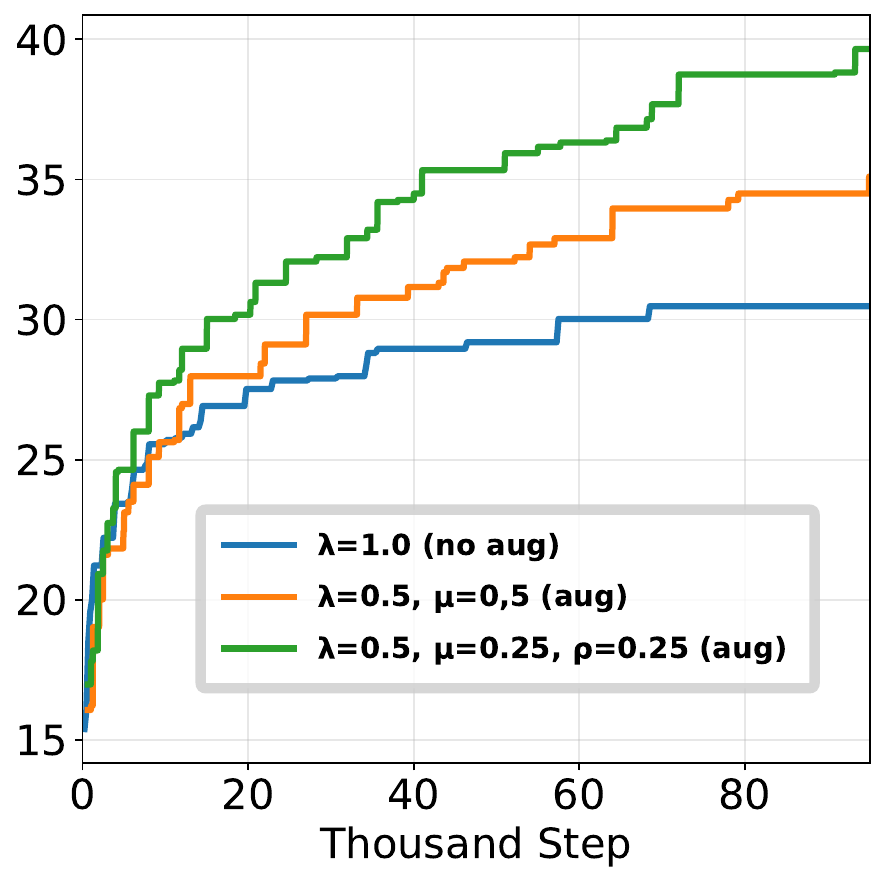}
    \caption{Evaluation Accuracy of Galactica-125M on GSM8K during RL training.}
    \label{fig:galactica_gsm_accuracy_main}
\end{figure*}

The three curves in Figure~\ref{fig:reward_curves} demonstrate that all three reward types increase steadily during training before eventually plateauing. These curves correspond to our semi-supervised setup utilizing Verifiable Reward (VR), Continuous CoT Reward (CCR), and Pretrained Reranker (PRR) on the combined GSM8K and GSM8K-aug datasets.

Furthermore, Figure~\ref{fig:galactica_gsm_accuracy_main} shows that semi-supervised RL, when combined with diverse reward signals (CCR and PRR), yields the highest overall performance.

Table~\ref{tab:galactica_gsm_tuning_coeff} presents an ablation study on the loss coefficients $\lambda$, $\mu$, and $\rho$.
It contains additional results for using the GSM8K as unsupervised data for combining VR and CRR rewards.

\begin{table*}[t]
\centering
\begin{tabular}{ccc|c c ccc c}
\toprule
\multicolumn{3}{c}{\textbf{Coefficients}} 
& \textbf{GSM8K} & \textbf{GSM-Plus} & \multicolumn{3}{c}{\textbf{GSM-Symbolic}} & \textbf{SVAMP} \\
\cmidrule(lr){1-3} \cmidrule(lr){6-8}
$\lambda$ & $\mu$ & $\rho$ & & & main & p1 & p2 & \\
\midrule
\multicolumn{9}{l}{\textbf{Reported by~\citet{trung2024reft}}} \\
\multicolumn{3}{l}{SFT\quad Complete} & 23.70 & --- & --- &--- &--- &--- \\
1.0 &  --  & --    & 29.80 & --- & ---& ---&--- &--- \\

\multicolumn{9}{l}{\textbf{Our reproduction}} \\
\multicolumn{3}{l}{SFT\quad Warm-up} & 21.83 & 12.49 & 20.24 & 4.88 & 1.80 & 29.50 \\
\multicolumn{3}{l}{SFT\quad Complete} & 25.25 & 15.14 & 22.88 & 7.40 & 3.64 & 32.30 \\
1.0 & 0.0 & 0.0 & 30.48 & 18.38 & 30.96 & 8.42 & 0.48 & 31.30 \\

\multicolumn{9}{l}{\textbf{GSM8K unsup}} \\
0.0 & 1.0 & 0.0 & 29.19 & 17.95 & 26.42 & 11.70 & {4.72} & 37.00 \\

\multicolumn{9}{l}{\textbf{GSM8K sup + GSM8K unsup}} \\
0.5 & 0.5 & 0.0 & 30.63 & 18.96 & 28.32 & 9.64 & 1.40 & 35.90 \\

\multicolumn{9}{l}{\textbf{Pseudo-labeled GSM8K-aug sup}} \\

\multicolumn{9}{l}{{Base 1}} \\
1.0 & 0.0 & 0.0 & 28.89 & 16.78 & 25.00 & 6.48 & 2.16 & 44.60 \\

\multicolumn{9}{l}{{Base 2}} \\
1.0 & 0.0 & 0.0 & 36.01 & 26.64 & 37.20 & 14.58 & 1.96 & 51.00 \\

\multicolumn{9}{l}{\textbf{GSM8K-aug unsup}} \\
0.0 & 1.0 & 0.0 & 33.13 & 21.28 & 30.38 & 11.36 & 3.32 & 44.70 \\
0.0 & 0.0 & 1.0 & 32.98 & 20.92 & 31.12 & 10.36 & 2.60 & 46.30 \\

0.0 & 0.5 & 0.5 & {40.71} & {25.33} & {37.74} & {14.62} & 3.00 & {50.70} \\

\multicolumn{9}{l}{\textbf{GSM8K sup + GSM8K-aug unsup}} \\

0.5 & 0.5 & 0.0 & 35.10 & 22.14 & 34.08 & 12.76 & 3.32 & 40.70 \\
0.5 & 0.0 & 0.5 & 39.27 & 25.04 & 36.86 & 12.84 & 1.44 & 43.30 \\
0.5 & 0.25 & 0.25 & {40.11} & {25.26} & {37.50} & {15.70} & 2.24 & 42.60 \\

\multicolumn{9}{l}{\textbf{GSM8K sup + GSM-Plus unsup}} \\
0.5 & 0.5 & 0.0 & 41.70 & 28.97 & 35.52 & 12.78 & 1.12 & 37.90 \\
0.5 & 0.25 & 0.25 & 44.66 & 30.67 & 38.74  & 10.88 & 0.84 & 32.50 \\
\bottomrule
\end{tabular}
\caption{Galactica-125M trained on GSM8K. Results with coefficients.}
\label{tab:galactica_gsm_tuning_coeff}
\end{table*}

\section{Implementation Details for Competition-level Math}
Experiments are performed on eight NVIDIA RTX Pro A6000 GPUs.

\subsection{Models and Datasets}
The student model is \texttt{Qwen/Qwen3-0.6B-Thinking}, and the CCR judge model is \texttt{Qwen/Qwen3.5-9B}. 

For training, we utilize the \texttt{di-zhang-fdu/AIME\_1983\_2024} and \texttt{BytedTsinghua-SIA/DAPO-Math-17k} datasets from Hugging Face. For evaluation, we use \texttt{math-ai/aime24}, \texttt{math-ai/aime25}, and \texttt{math-ai/aime26}.

\subsection{Hyperparameters}
\paragraph{CCR Reward.} We set $\tau=0.5$ and $a=0.1$.
\paragraph{GRPO Settings.} We employ the Group Relative Policy Optimization (GRPO) algorithm via the TRL and vLLM libraries. Across all experiments, we set the KL penalty coefficient $\beta=0$ and the group size $G=7$. In our semi-supervised setup, we leverage TRL's ability to assign sampling probabilities to different reward formulations, utilizing a $\lambda=0.1$ weight for supervised data and a $\mu=0.9$ weight for unsupervised data. The model is trained for approximately one epoch. Table \ref{apdx:aime_tune} contains an ablation study on the loss coefficients $\lambda$ and $\rho$ in a semi-supervised setup.

\begin{table*}[t]
\centering
\begin{tabular}{c c c c c}
\toprule
\textbf{$\lambda$}
& \textbf{$\mu$}
& \textbf{AIME'24}
& \textbf{AIME'25}
& \textbf{AIME'26} \\
\midrule
0.1 & 0.9 & \textbf{16.67} & \textbf{20.73} & \textbf{15.21} \\
0.5 & 0.5 & 11.25 & 16.88 & 11.87 \\
0.9 & 0.1 & 15.42 & 19.06 & 14.90 \\
\bottomrule
\end{tabular}
\caption{Avg@32 results of Qwen3-0.6B-thinking on AIME with semi-supervised RL.}
\label{apdx:aime_tune}
\end{table*}

\paragraph{Optimization and System Details.}
To accelerate training, we apply LoRA fine-tuning targeting \texttt{all-linear} modules with rank $r=8$ and scaling factor $\alpha=16$. We use a global batch size of 56, distributed across 7 GPUs with 8 gradient accumulation steps and a local batch size of 1. The learning rate is set to $5\times 10^{-5}$. We dedicate a single GPU exclusively for vLLM inference. The maximum number of completion tokens is set to 20,000 in TRL and 21,600 in vLLM.
\section{Model Collapse in Unsupervised Training}
\label{sec:hacking}
\begin{figure*}[htbp]
    \centering
    \includegraphics[width=0.5\linewidth]{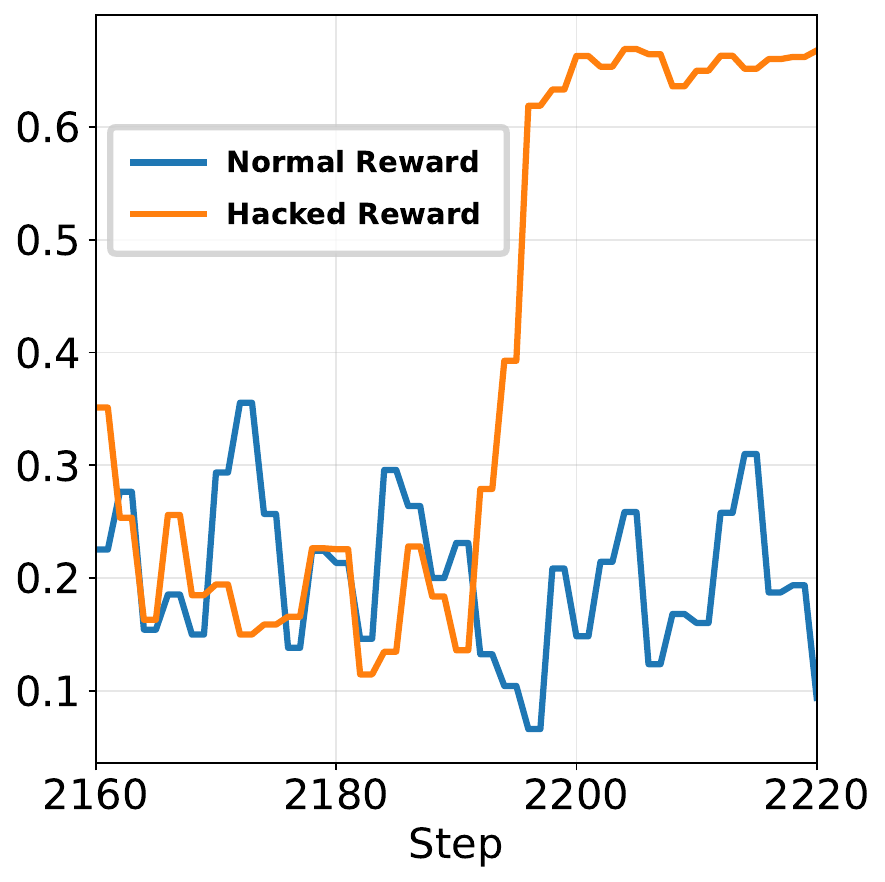}
    \caption{Judge Mean Reward Curve}
    \label{fig:hacked_curve}
\end{figure*}

Reward hacking has been reported in existing RL literature~\citep{uehara2024ctdiffusioncontrol,skalse2025rewardhacking}. It is a phenomenon where the nominal (or proxy) reward values are high (e.g., the aesthetic score of a generated image, the plausibility of a generated code snippet), while the generation itself does not make sense. In our experiments, this may occur without VR. For example, as seen in Figure~\ref{fig:hacked_example}, the hacked response does not generate any code (or in some cases, generates a number with no reasoning whatsoever and returns directly), but the reward is much higher than the incorrect response with reasonable code. As we further demonstrate in Figure~\ref{fig:hacked_curve}, the transition from a stable state to a failure state happens quickly. \textbf{Empirically, this phenomenon can be avoided by using VR together with CCR, showing the strength of our semi-supervised setup.} Although training without ground-truth labels is not the focus of this work, below we show some strategies to make \textit{unsupervised} training more stable.

\paragraph{Mitigation Strategies}
We present some intuitive mitigation strategies when CCR is the only available reward signal (fully unsupervised), but more rigorous treatments are left for future exploration. Occasionally, reward hacking persists on the unsupervised dataset even when the supervised dataset yields reasonable signals. This may lead to model collapse or drastically slow down training. To alleviate this, we (1) explicitly instruct the small LMs at the end of the docstring: ``\texttt{Please provide full reasoning before computing the result; do not answer with the result only.}'' and (2) apply a mask so that if a generated code snippet is fewer than 4 lines, we assign a judge reward of 0. This modification is only applied to the GSM8K sup + GSM-Plus unsup setting using CodeGen in Table~\ref{tab:codegen_gsm_results}. However, these patches may not fully resolve the issue, since our modifications occasionally led to consistent zero rewards and, consequently, model collapse. They also cause performance degradation in runs that do not exhibit reward hacking.

The instance of reward hacking reported in~\citet{trung2024reft} on the MathQA-Multiple Choice dataset, where the model reports the correct answer option despite using incorrect reasoning, is different from our encounter, where the model provides little to no response without any reasoning. Nevertheless, both scenarios achieve high rewards with low-quality reasoning. Future work on new regularization methods is therefore necessary.

\begin{figure*}[htbp]
\centering
\begin{tcolorbox}[
  colback=gray!10,
  colframe=black,
  boxrule=0.4pt,
  arc=2mm,
  width=\textwidth
]
\small
\textbf{Question:} each pack of dvds costs 76 dollars. if there is a discount of 25 dollars on each pack how much do you have to pay to buy each pack?
\end{tcolorbox}
\begin{minipage}{0.32\textwidth}
\begin{tcolorbox}[title=\textbf{Correct Response}, halign title=flush center, equal height group=grp]
\tiny
\begin{verbatim}
def solution():
    """each pack of dvds costs 76...
    ...how much do you have to pay?"""
    cost_per_pack = 76
    discount_cost = 25
    total_cost = cost_per_pack 
                - discount_cost
    return total_cost
\end{verbatim}
\tcblower
\centering \textbf{Reward: 0.728}
\end{tcolorbox}
\end{minipage}
\begin{minipage}{0.33\textwidth}
\begin{tcolorbox}[title=\textbf{Incorrect Response}, halign title=flush center, equal height group=grp]
\tiny
\begin{verbatim}
def solution():
    """each pack of dvds costs 76...
    ...how much do you have to pay?"""
    cost_per_pack = 76
    discount_amount = 25
    total_cost = cost_per_pack 
                * discount_amount
    return total_cost
\end{verbatim}
\tcblower
\centering \textbf{Reward: 0.303 (clipped to 0)}
\end{tcolorbox}
\end{minipage}
\begin{minipage}{0.33\textwidth}
\begin{tcolorbox}[title=\textbf{Hacked Response}, halign title=flush center, equal height group=grp]
\tiny
\begin{verbatim}
def solution():
    """each pack of dvds costs 76...
    ...how much do you have to pay?"""
\end{verbatim}
\vspace{4.85em}
\tcblower
\centering \textbf{Reward: 0.635}
\end{tcolorbox}
\end{minipage}

\caption{Example completions with CCR scores only. Notice how the hacked response does not generate anything given the declaration and docstring, yet obtaining a good score.
}
\label{fig:hacked_example}
\end{figure*}


\end{document}